\documentclass[10pt, conference]{IEEEtran}

\usepackage{cite}
\usepackage{url}
\usepackage{amsmath}
\usepackage{amssymb}
\usepackage{bbm}
\usepackage{enumitem}
\usepackage{graphicx}
\DeclareGraphicsExtensions{.pdf,.jpeg,.png}

\title{\LARGE Learning to Associate Words and Images\\
Using a Large-scale Graph}

\author{\IEEEauthorblockN{Heqing Ya, Haonan Sun, Jeffrey Helt, Tai Sing Lee}
\IEEEauthorblockA{School of Computer Science\\
Carnegie Mellon University, Pittsburgh, PA 15213\\
\texttt{\{heqingy@cs,haonans@cs,jhelt@cs,tai@cnbc\}.cmu.edu}}}

\begin{document}

\maketitle

\begin{abstract}
We develop an approach for unsupervised learning of associations between co-occurring perceptual events using a large graph. We applied this approach to successfully solve the image captcha of China's railroad system. The approach is based on the principle of suspicious coincidence, originally proposed by Barlow \cite{Barlow1987Cerebral}, who argued that the brain builds a statistical model of the world by learning associations between events that repeatedly co-occur. In this particular problem, a user is presented with a deformed picture of a Chinese phrase and eight low-resolution images. They must quickly select the relevant images in order to purchase their train tickets. This problem presents several challenges: (1) the teaching labels for both the Chinese phrases and the images were not available for supervised learning, (2) no pre-trained deep convolutional neural networks are available for recognizing these Chinese phrases or the presented images, and (3) each captcha must be solved within a few seconds. We collected 2.6 million captchas, with 2.6 million deformed Chinese phrases and over 21 million images. From these data, we constructed an association graph, composed of over 6 million vertices, and linked these vertices based on co-occurrence information and feature similarity between pairs of images. We then trained a deep convolutional neural network to learn a projection of the Chinese phrases onto a 230-dimensional latent space. Using label propagation, we computed the likelihood of each of the eight images conditioned on the latent space projection of the deformed phrase for each captcha. The resulting system solved captchas with 77\% accuracy in 2 seconds on average. Our work, in answering this practical challenge, illustrates the power of this class of unsupervised association learning techniques, which may be related to the brain's general strategy for associating language stimuli with visual objects on the principle of suspicious coincidence.
\end{abstract}


\begin{figure}
\centering
\includegraphics[width=\linewidth]{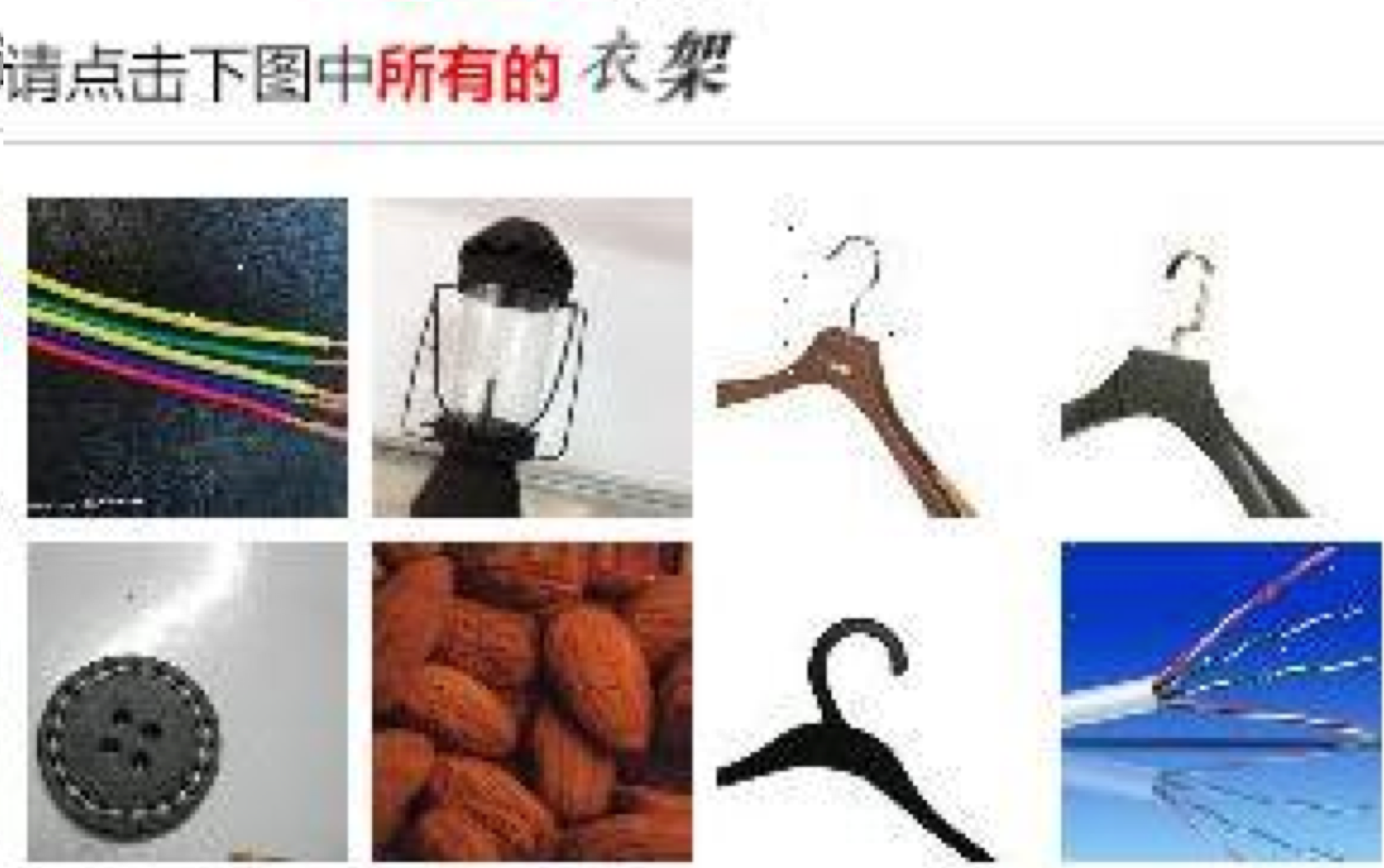}
\caption{Example captcha. The phrase translates to ``hangers.'' The characters of interest are to the right of those in red.}
\label{fig:captcha}
\end{figure}

\section{Introduction}

As the railway operator for a country of over one billion people, China Railway is one of world's largest, selling 4.5 million tickets per day \cite{Zhu2016China}. During peak holiday travel periods, this rate can increase to as much as 15 thousand tickets per minute. Tickets for China Railway trains can be purchased through their website, \url{www.12306.cn}. Due to the high demand of tickets, their website is a primary target for sophisticated scalpers, who use automated attacks to purchase large volumes of tickets in order to resell them for a significant profit. Thus, it is critical that China Railway employ a strong captcha \footnote{We use the lowercase acronym here for improved readability.} system to protect China's citizens from being swindled. In response to these threats, China Railway created a captcha that is similar to Google's reCaptcha \cite{Shet2014Are}. When attempting to purchase a ticket, a user is presented with a phrase, written in distorted Chinese characters, and eight images. The user must select all of the images that are relevant to the specified phrase. For an example, see Figure \ref{fig:captcha}.

There are a few significant aspects of this system that make this task particularly difficult for automated agents. First, unlike the other text in the captcha, the relevant phrase is actually a distorted, low-resolution image of Chinese characters. Henceforth, we will refer to the images of the phrases as codes in order to avoid confusion with the other images presented in the captcha. In addition to the codes, the presented images are also of low quality and subject to random noise. The low quality of both types of images makes it difficult to develop effective algorithms. Finally, and most importantly, there is a complete absence of labeled data for both the codes and the images. This severely restricted our ability to leverage recent advancements in supervised learning that have become popular in recent years \cite{Krizhevsky2012ImageNet}. Thus, we turn to an unsupervised, graph-based approach to solve this problem.

For an automated system to defeat this captcha, it must learn to associate the codes and their corresponding images. In formulating our approach, we made the following assumptions. First, when a code is presented, we assume that there is at least one relevant image. Second, we assume there may be more than one relevant image but irrelevant images tend to appear only once and at random. Based on these assumptions, we were able to exploit two kinds of co-occurrence relationships for this task: between pairs of images and between codes and images, based on their co-occurrence in each captcha.

We see a few distinct subproblems of this task. Our approach needs to learn to map variations of a single code to the same point in latent space. Second, we need to cluster images of the same type of object. Finally, we need to learn the correct association between the latent representations of the codes and the image clusters. By exploiting the co-occurrence information, we can couple the learning and inference of these processes using a graph.

Our goal was to construct a large graph to relate both images to each other as well as the codes and the images. To do this, we first trained a deep convolutional neural network (DCNN) to develop feature representations for Chinese characters that were robust to a variety of deformations and transformations. We randomly selected 230 Chinese phrases, including some ground truth phrases, and generated thousands of random transformations of each of these phrases to use as our training set. Using this DCNN, we then projected all of the codes into a probability distribution in this 230-dimensional ``latent'' space. Note that the exact choice of the 230 phrases was unimportant because our goal was not to classify the codes; we simply wanted a useful latent space in which to cluster them. However, we did include some ground truth phrases in this set of 230 in order to evaluate this DCNN component in isolation, as will be discussed in Section \ref{sec:codeRecognition}.

\begin{figure}
\centering
\includegraphics[width=\linewidth]{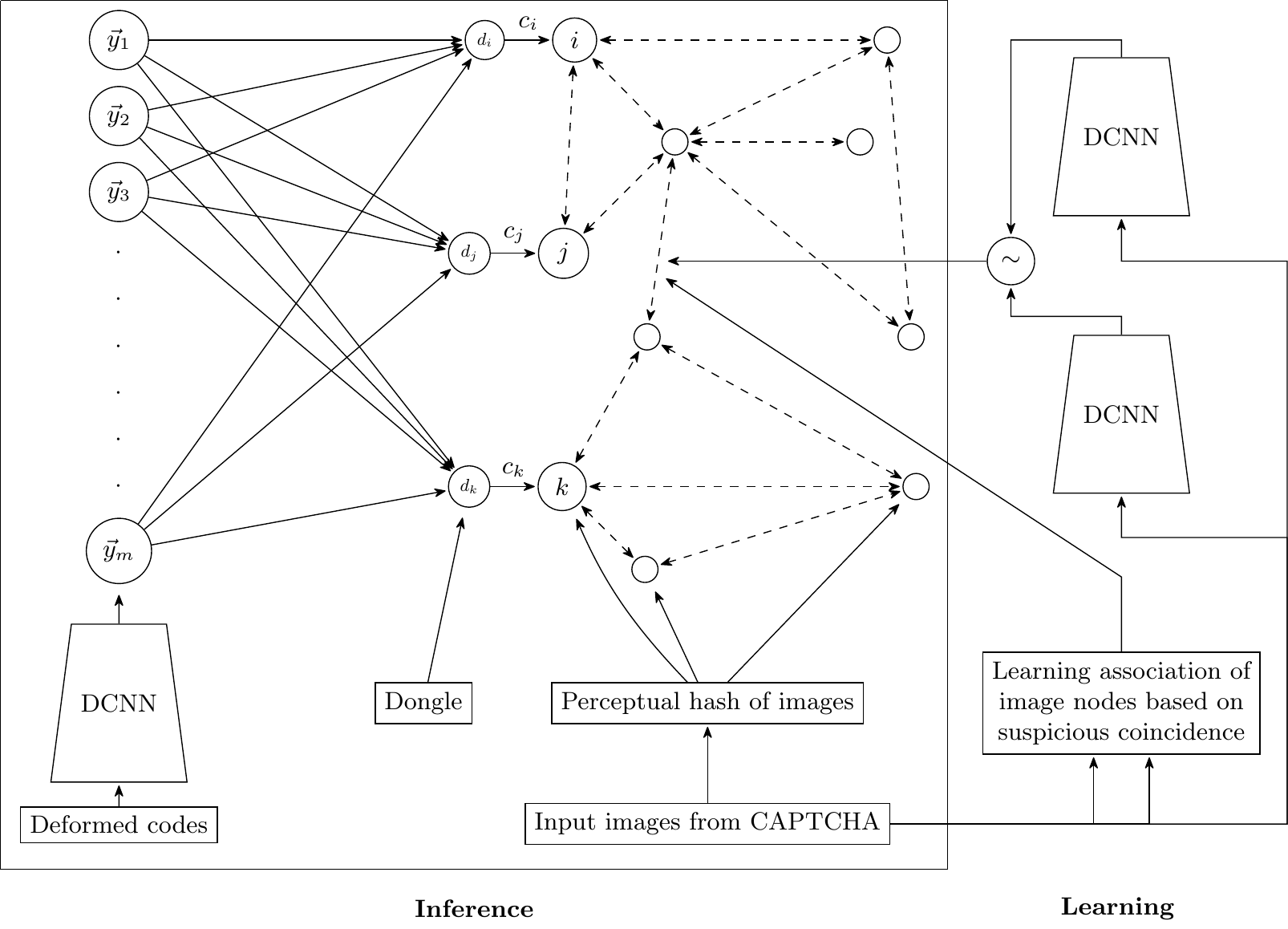}
\caption{A graphical outline of the learning and inference phases. The code vertices are shown on the left, labeled with their latent representations $\vec{y}$. They are connected to the dongles of image vertices $i$, $j$, and $k$. $c_i$ is the number of captchas in which image $i$ appeared. For clarity, the dongles and connections of other image vertices have been omitted. During inference, only the graph inside the rectangle is used. During unsupervised learning, the associations between image vertices are learned based on the co-occurrence of the input images and the cosine similarity of the DCNN feature vectors for each image.}
\label{fig:graph}

\end{figure}

Next, we constructed a weighted graph of 3.5 million images. In the graph, each vertex was associated with an image, and we connected pairs of image vertices based on feature similarity and their co-occurrence statistics. These vertices were further linked to an additional set of vertices, one for each of the $m = \text{2.6 million}$ codes, via a dongle. The dongle was used to encode the number of times an image appeared in a captcha and thus our confidence in the relationship between the images and codes. Associated with each of these code vertices was its 230-dimensional latent representation, denoted $\vec{y}$. For each of the code vertices, we connected it to the dongle of an image vertex if the code and the image co-occurred in some captcha. Finally, we performed label propagation on this graph, as described in \cite{Zhu2003Semi}, to learn a latent representation for each of the images. Armed with the representations of the codes and images in the same latent space, we then chose the relevant images given a presented code. A representative subsection of the resulting graph is shown in Figure \ref{fig:graph}, and for more details, see Section \ref{sec:graphConstruction}.

Our results are promising. Our system solved 77\% of test image challenges with an average response time of around 2 seconds. This is especially impressive because findings show that average human performance is between 61\% and 98\% on image-based captchas depending on the specific captcha implementation \cite{Bursztein2010How}. \footnote{Reports from the China News Service and China Economics Net suggest that this particular captcha is especially difficult. They state that only 6\%, 27\%, and 65\% of people successfully pass the captcha on their first, second, and third attempts, respectively.} In addition, there is promise of improvement over a recently released system by \cite{Sivakorn2016I}. Their model solved about 71\% of Google's image reCaptcha challenges \cite{Shet2014Are}, although direct comparisons of the two systems are difficult due to differences in the captcha implementations.

The major contributions of this work are as follows:
\begin{enumerate}[wide]
\item We designed a novel graph-based approach for learning the associations between textual information and images. We leveraged a number of different techniques to learn these associations in an unsupervised setting. We demonstrated the effectiveness of our approach on an image captcha implementation and show promising results.

\item We demonstrated that the exploitation of contextual co-occurrence information and the transfer of deep neural network features are powerful techniques in the absence of labeled data.
\end{enumerate}

\section{Background \& Related Work}

Following its inception in 2003 \cite{Ahn2003CAPTCHA}, companies have employed captchas to secure their websites from spammers and bots. Due to the challenge they present to automated systems, captchas have long been used a benchmark for artificial intelligence and computer vision communities. Thus, a symbiotic relationship formed. As research advanced in these communities, captcha developers were forced to innovate, creating new, more difficult benchmarks. Rapid advances in text and character recognition have rendered the first two iterations of text-based captcha systems obsolete \cite{Goodfellow2013Multi,2013Vicarious}. In fact, a recent paper has demonstrated the effectiveness of a generic text-based captcha solver that can be applied to new implementations without implementation-specific modifications \cite{Bursztein2014End}. In response, large companies such as Google have introduced image-based systems \cite{Shet2014Are} where a user is presented with a set of images and an accompanying statement or question. Based on the contents of the statement or question, the user must then select the correct images to prove that he or she is not a robot. However, as advancements continue in the computer vision and artificial intelligence communities, these new systems are now being threatened.

Recently, Sivakorn et al.\ \cite{Sivakorn2016I} detailed an approach that achieved roughly 71\% accuracy on Google's reCaptcha service and 83\% on Facebook's image-based system. However, our work differs drastically from theirs. They leverage the power of online image annotation services and Google's own reverse image search to train DCNNs. We found that the number of images used in the China Railway's captcha, roughly 21 million, made online annotation services prohibitively expensive. Similarly, they employ supervised learning techniques to recognize the presented English text. In our case, the phrases were more difficult to classify since they were images of Chinese characters. This is both because of the content and the deformations. We are not aware of any pre-trained models for recognizing Chinese characters that could overcome the level of noise introduced by this captcha system.

Over the past few years, computer vision research has been revolutionized by the use of DCNNs. These techniques were popularized by Krizhevsky et al.\ \cite{Krizhevsky2012ImageNet} for image classification, and subsequently, have been successfully applied to other computer vision tasks, including video classification \cite{Karpathy2014Large} and action recognition \cite{Simonyan2014Two}. However, training a DCNN requires massive amounts of labeled data. Luckily, recent findings have shown that feature representations extracted from pre-trained DCNNs can be effectively leveraged for other tasks. In particular, Razavian et al.\ \cite{Razavian2014CNN} showed that these representations, in combination with a linear SVM classifier, outperform state-of-the-art systems for a variety of tasks. More recently, work has shown that DCNN feature representations transfer well to similarity perception tasks \cite{Lin2016Transfer}. Our results lend credence to these findings. Although our paper is not focused on transfer learning, we leverage the power of extracted feature representations, in combination with co-occurrence statistics, to approximate the manifold of images in the graph. Perhaps even more interesting is that our approach was still effective despite using a sparse feature representation for each image (keeping only the largest 10\% of the weights).

Techniques for incorporating contextual information have long been used in the speech recognition community through the use of language models. However, less work has been done to learn from contextual information for computer vision tasks. Recently, Bengio et al.\ \cite{Bengio2013Using} have shown improvements in object recognition by incorporating label co-occurrence statistics mined from web pages. They achieve significant improvement over a baseline classifier, similar to AlexNet \cite{Krizhevsky2012ImageNet}, by optimizing an objective function that is a relaxation of the traditional Ising model \cite{Ising1925Beitrag} to allow real-valued states. As noted in \cite{Zhu2005Semi}, the label propagation algorithm also optimizes a relaxed Ising objective function.

Although our work bears strong resemblance to that of \cite{Bengio2013Using}, there are key differences. First, they derive similarity scores for each pair of labels based on term proximity in text-based web data. However, since images present our primary challenge, we instead derive the similarity between pairs of images based on their feature representations and co-occurrence statistics. Second, they use a pre-trained, external classifier to initialize a vector of object scores for each image. We, on the other hand, leverage image-code co-occurrence statistics. Although it is difficult to compare the performance of these models due to differences in the data set, we believe our results show promise of improvement.

Multi-modal associative learning has a enormous literature, and we do not aim to cover it all in this paper. However, we will note two important connections to our work. The first is \cite{Hinton2006Fast}, in which Hinton et al.\ showed that a model could be trained to learn associations between images and text using a fully connected association layer. They showed that their generative model was able to outperform discriminative methods at the time. However, this work was demonstrated on the the well-known MNIST database consisting of labeled images of hand-written digits. More recently, Mansimov et al.\ \cite{Mansimov2015Generating} leveraged recent advances in variational autoencoders to train a generative model to learn associations between images and their captions. Although our approach also aims to learn associations between text and images, we faced a unique set of challenges in defeating this captcha. In our case, we did not know the exact correspondence between the phrases and the images, and thus, we were unable to use this information as a training signal. Thus, rather than learning one, our model uses a fixed association mechanism, the co-occurrence between codes and images as a training signal. In addition, due to the noisiness and content of the Chinese phrases, it was impossible to leverage useful latent representations of the codes from a pre-existing model, as was done for the captions in \cite{Mansimov2015Generating}.

\section{Proposed Approach}

We will start by describing our approach to learning a latent representation for the codes in Section \ref{sec:latentRepresentation}. Then, we will describe in detail the construction of the code and image graph in Section \ref{sec:graphConstruction}. We will conclude with descriptions of the label propagation algorithm and the process by which our system selects the relevant images in a captcha in Sections \ref{sec:labelPropagation} and \ref{sec:imageSelection}, respectively.

\subsection{Learning a Latent Representation} \label{sec:latentRepresentation}

The phrase in each captcha is a low-resolution image of Chinese characters that is distorted by random noise. Unfortunately, we found that pre-trained CNN models did not produce useful feature representations of the codes. We suspect that this was caused by two issues: the uniqueness of the content and the significant noise. Chinese characters comprise a unique class of images and are unlikely to be similar to any of the images in the training set of a pre-trained CNN. To make matters worse, the noise introduced by the captcha is a combination of wave-like distortions along the vertical and horizontal axes. In addition, the system introduced random ASCII characters to the images to serve as distractors. Thus, instead of simply classifying the codes using a pre-trained model, we sought to train a model that would cluster them in a useful ``latent'' space where known, random phrases were used as basis functions. For this task, we trained a LeNet-5 \cite{Lecun1998Gradient} DCNN as follows:

In order to define a useful latent space in which to represent the codes, we selected a random set of Chinese phrases. Some of these phrases were translations of the codes and others were randomly selected from an external dictionary. In total, we assembled a set of 230 phrases to serve as the basis for our latent representation. We reiterate that the exact choice of these phrases is unimportant because we did not seek to classify the codes. Instead, we wished to cluster them in this latent space. In addition, we did not assemble an exhaustive set of translations of the codes. In other words, there were many codes whose true translation is not among the set of 230.

\begin{figure}
\centering
\includegraphics[width=\linewidth]{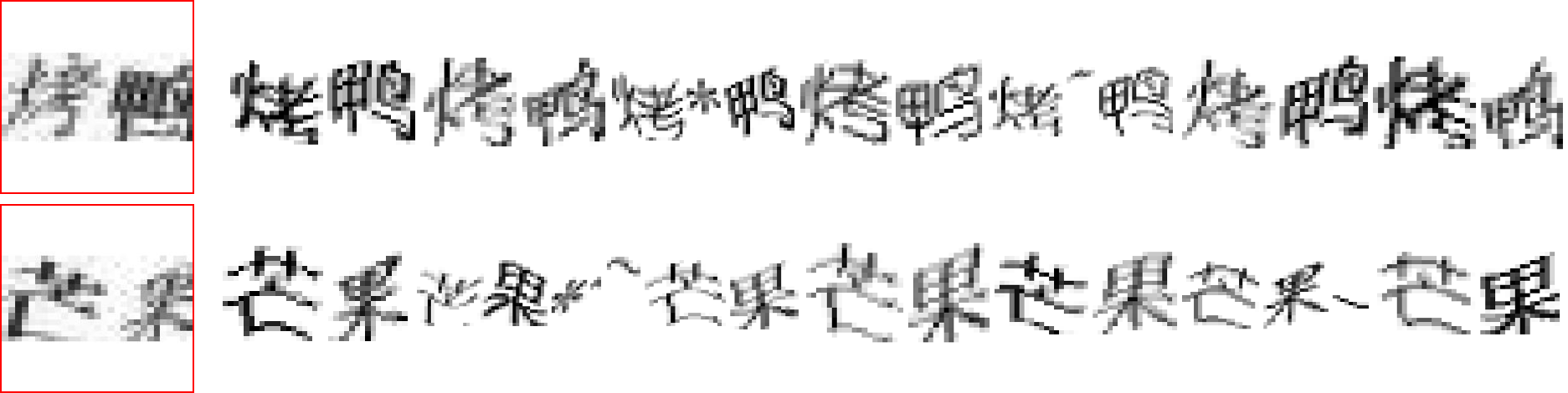}
\caption{Real codes next to generated mock codes. The real codes are circled in red. The code in the first row translates to ``roast duck,'' and the code in the second row translates to ``mango.''}
\label{fig:mockLabels}
\end{figure}

Due to the lack of labeled data, we built a mock code generator to produce mock data for the 230 phrases. The generator approximated the noise observed in the real codes using sine and cosine transformations along the vertical and horizontal axes as well as randomly inserted ASCII symbols to simulate the noise in the real codes. See Figure \ref{fig:mockLabels} for examples of generated codes. For each class, we generated 1,000 mock codes. We combined this mock data set with hand-labeled codes from 24,000 captchas to produce a training set of 254,000 data.

We first used the mock data to train the DCNN. This pre-training allowed the network to learn invariant convolutional filters specifically for Chinese characters. We then fixed these layers and fine-tuned the model using the hand-labeled data. The idea here was to then cluster the codes in the latent space.

\subsection{Assembling the Graph} \label{sec:graphConstruction}

Using an automated crawler, we collected 2,637,184 captchas from \url{www.12306.cn}, resulting in 21,097,472 unlabeled images. The lack of labels ruled out the possibility of using supervised learning techniques, and the low quality of the images prevented us from simply classifying the images using a pre-trained model. However, we suspected that clustering the images could provide valuable information for defeating the captcha. Thus, we used an unsupervised, graph-based approach to solve this problem.

\emph{Perceptual Hashing:} We aimed to build a graph of all of the images in our database. Furthermore, when our system processes a displayed captcha, it must be able to compare the 8 images in the captcha with the vertices in our graph. All of this must be completed within 2 seconds, a daunting task. To achieve this, we employed an existing technique called perceptual hashing to identify duplicate images and create compact representations of each image. Unlike cryptographic hash functions, perceptual hash functions are robust to noise \cite{Klinger2010pHash}. In other words, two inputs that are similar will have similar, if not the same, perceptual hash values. This feature was extremely useful for grouping duplicates despite the pixel-level impulse noise applied to the images. By processing our image database with a perceptual hash function, we were able to reduce our original set of approximately 21 million images to a set of about 3.5 million.

\begin{figure}
\includegraphics[width=\linewidth]{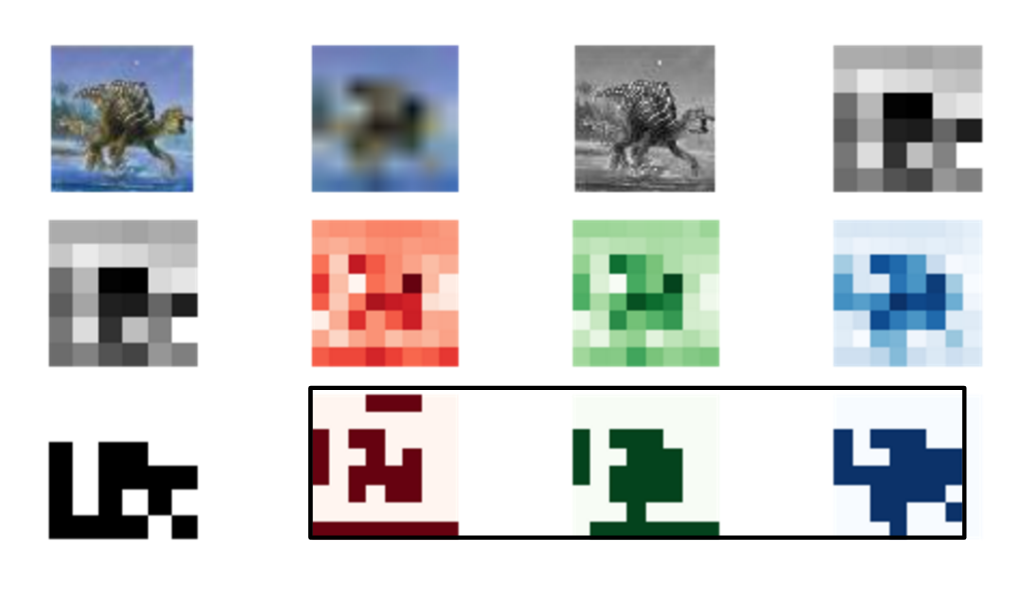}
\caption{Perceptual hashes of an image. In the top row, we see the image, in both color and grayscale, next to downsampled versions. In the middle row, we separate the downsampled image into its three color channels. In the bottom row, we see the binary version of the grayscale image and each of the three channels. Note that the grayscale image is not used in the final vector representation of each image.}
\label{fig:phash}
\end{figure}

The perceptual hash values also provided a method for fast retrieval of images. For each image, we calculated its perceptual hash value as follows. For each of the three RGB channels, we downsampled the original image to produce an $8 \times 8$ representation. Next, we normalized the 64 pixel values and rounded each of them. The resulting image has binary pixel values. Finally, we combined the results into a 192-bit vector representation of the image. For an illustrative example, see Figure \ref{fig:phash}. The perceptual hash value of any image was quick and easy to compute, and comparisons between values were also fast. To further increase the responsiveness of our system, we bucketed images based on the perceptual hash value of their grayscale image to reduce the number of necessary comparisons.

\begin{figure}
\centering
\includegraphics[width=\linewidth]{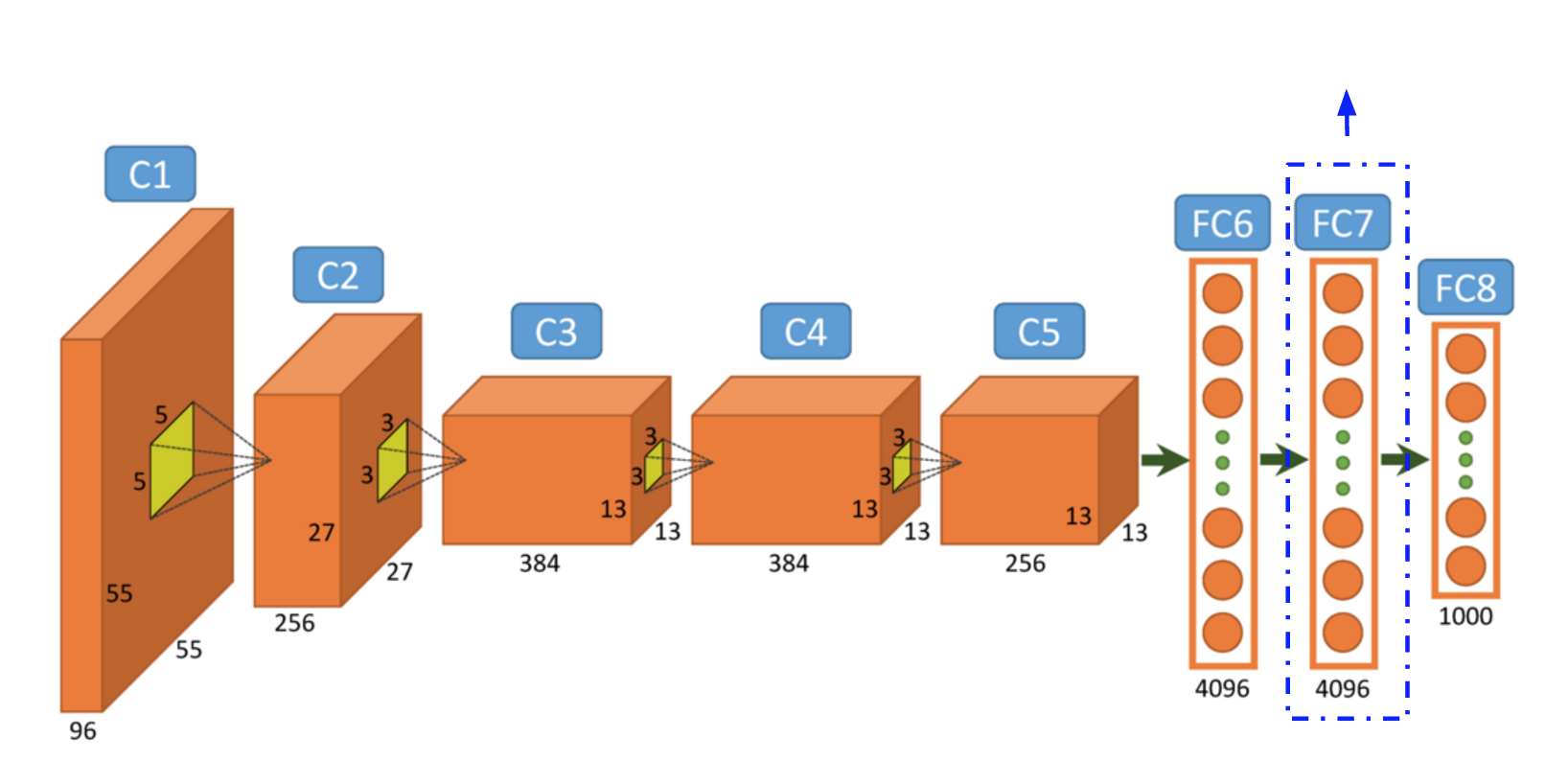}
\caption{CaffeNet. Each layer is labeled with a blue box. We extracted the feature representation of each image at FC7 layer. Image courtesy of \cite{Hu2015Transferring}.}
\label{fig:caffenet}
\end{figure}

\emph{Similarity Measure:} Extracted DCNN feature vectors have proven to be powerful representations that transfer well to new tasks \cite{Razavian2014CNN,Lin2016Transfer}. Thus, we decided to use these feature representations as a basis for a pairwise similarity measure for the images in our data set. We used a pre-trained Caffe implementation of AlexNet \cite{Krizhevsky2012ImageNet} called ``CaffeNet.'' A visual representation of the network is shown in Figure \ref{fig:caffenet}. For each image, we extracted the FC7 (fully-connected 7) layer feature representation. However, since even our reduced data set was still very large, we reduced the storage requirements of the feature representations by keeping only the largest 10\% of the weights, reducing the original 4096-dimensional representation to a sparse vector.

In addition to calculating the cosine similarity of the feature representations for each pair of images, we also mined our database of 2,637,184 captchas to count the number of times each pair of images co-occurred. Based on both of these sources of information, we then wanted to approximate the manifold structure of the images by calculating a pairwise similarity measure. Thus, we propose the following function for each pair of images $i, j$:
\[
\text{sim}(i,j) = \left(\frac{\vec{f_i} \cdot \vec{f_j}}{|\vec{f_i}|\cdot|\vec{f_j}|}\right) \cdot g(c_{ij})
\]
where $\vec{f_i}$ is the sparse feature vector extracted from CaffeNet's FC7 layer for image $i$, and $c_{ij}$ is the number of co-occurrences of images $i$ and $j$. $g(x) := x^2$ is a re-scale function, which we used to adjust the gap between small and large co-occurrence values. This similarity measure was an effective way to combine the high-level feature representations with the co-occurrence information. We believe this worked well because images of the same type of object were likely to co-occur in captchas where they were relevant. However, irrelevant images were shown at random. Thus, similar images (i.e.\ those containing similar objects) had higher similarities, and the random noise introduced by distractors canceled out.

\emph{Graph Construction:} We constructed a graph in three steps, which we will now describe in turn. A representative subsection of the final result can be seen in Figure \ref{fig:graph}. We started by constructing a fully-connected, directed graph $G = (U, E)$ where each $u \in U$ corresponds to an image. We denote the set of vertices by $U$ to emphasize that they are unlabeled. Associated with each edge is a weight, which is equal to the similarity (as defined above) between the two images, that is, $w_{ij} = \text{sim}(i,j)$. Note that because most pairs of images do not co-occur in any captcha, this weighted graph is very sparse.

We then augment each vertex in the graph with a ``dongle,'' as described in \cite{Zhu2003Semi}. Each dongle has only one outgoing edge, from it to its corresponding image vertex. The weight of these edges is equal to the total number of captchas in which each image was shown. Since a dongle pertains to exactly one image and does not have an initial distribution over the latent basis functions, we add these dongles to $U$. In implementation, we treated a dongle and its corresponding image vertex as a single vertex.

Finally, we created a vertex $l \in L$ for each code. Associated with each of these vertices is the probability vector $\vec{y}$ over the 230 basis functions. As a result, we denote this set $L$ to emphasize that these are the ``labeled'' vertices in our graph. This adheres to the original label propagation terminology. We connected these code vertices to the rest of the graph using the image-code co-occurrence information. For every $l \in L$, we add an edge from it to a dongle if the code co-occurred with the dongle's corresponding image in some captcha. For simplicity, we assume that all codes are unique, so as a result, each vertex $l$ will have 8 outgoing edges. The weight of each of these edges is equal to 1.

The intuition behind this construction is as follows. The information from the probability vectors associated with each code will be used to inform the distributions of the images that co-occurred with that code. These are combined to form a prior distribution over the basis functions for each of the images. The edges between the dongles and the image vertices reflect how confident we are in that prior. We are much more confident in the prior distribution of an image that we saw 100 times than we are in the that of an image we only saw a few times. Finally, the weights between the image vertices approximates the manifold structure of the images. This provides valuable information as we try to learn the correct associations between images and codes.

An alternative interpretation of this graph structure is to remove the code vertices altogether and instead associate a probability vector with each of the dongles. Recall that each dongle is associated with exactly one image. To calculate the probability vector $\vec{y}_i$ for each of the dongles, we take an average of the probability vectors of each of the codes with which the dongle's associated image co-occurred. In other words, let $i$ be the index of the image associated the dongle and $m$ be the size of $L$. Then,
\[
	\vec{y}_i = \alpha^i_1 \vec{y}_1 + \ldots + \alpha^i_m \vec{y}_m.
\]
where each $\alpha^i_k$ is obtained using co-occurrence information between that code and the image:
\[
	\alpha^i_k = \frac{\mathbbm{1}\left\{A_{ik}\right\}}{\sum_k \mathbbm{1}\left\{A_{ik}\right\}}
\]
where $A_{ik}$ is the event that code $k$ co-occurred with image $i$. Note that these averages will be calculated over a very small fraction of the codes. Using this formulation, our approach can be seen as an unsupervised version of the selective attention or alignment mechanisms presented in  \cite{Bahdanau2014Neural,Bahdanau2015End,Gregor2015DRAW,Mansimov2015Generating,Karpathy2015Deep}.

\subsection{Label Propagation} \label{sec:labelPropagation}
We can now combine the knowledge gained from clustering the codes and the approximate manifold structure of the images. To do this, we implemented the label propagation algorithm proposed by Zhu et al.\ in \cite{Zhu2003Semi}. Let $p$ denote the dimension of our latent representation. In our case, $p=230$. Given the graph of unlabeled vertices $U$, labeled vertices $L$, and weighted edges $E$, the label propagation algorithm computes a vector-valued function $f : V \to \mathbb{R}^p$ where $V = U \cup L$. $f$ is constrained such that for all $l \in L$, $f(i) = f_l(i) = \vec{y}_i$. In other words, we clamp the output of $f$ for all of the labeled vertices such that the output is equal to their latent representations obtained from the DCNN. In our case, the code vertices in the graph serve as our ``labeled'' set.

Given this construction, the algorithm aims to minimize the following energy function:
\[
	E(Y)=\frac{1}{2}\sum_{i,j}w_{ij}(\vec{y}_i - \vec{y}_j)^2
\]
where each of entry of $Y$, $y_{ij}$, is the activation of basis function $j$ for image $i$. $f(i) = \vec{y}_i$ denotes the $i$th row of $Y$. Note that $Y \in \mathbb{R}^{n \times p}$ where $n = |U| + |L|$. For unlabeled vertices in the graph, we randomly initialize $\vec{y}$. Intuitively, this energy function seeks a function $f$ such that the outputs of $f$ for two vertices $i$ and $j$ are similar if the weight on the edge between them is large. In our case, since edges between the image vertices are weighted by their similarity, the outputs of nearby images will have similar latent representations. As shown in \cite{Zhu2003Semi}, the optimal solution for $Y$ occurs when the probability distributions for unlabeled vertices are equal to a weighted average of the distributions of their neighbors. In other words,
\[
	\vec{y}_j=\frac{\sum_i w_{ij}\vec{y}_i}{\sum_i w_{ij}} \text{, for } j \in U
\]
Note that we do not update the distributions of the code vertices because their values are clamped. The intuition here is that we want to push the distributions  from the code vertices through the dongles to the images with which they co-occurred. The label propagation algorithm will then push the influence of the distributions through the high density areas of the image manifold. This allows us to learn from both the image manifold structure as well as the codes' latent representations.

In implementation, we iterate over the unlabeled vertices and update their distributions with the weighted average of the distributions of its neighbors. We repeat this process until the distributions converge.

\subsection{Image Selection} \label{sec:imageSelection}
When presented with a captcha, our algorithm must select the subset of the images that correspond to the presented code. Our approach here is simple. We first use the perceptual hash of each image in the captcha to lookup its corresponding vertex in the graph. Then, we select those images whose largest activation among the 230 components of the latent representation matches that of the code. The rules for selecting the images can be summarized as follows:
\begin{enumerate}
\item Select at most four images whose largest activation matches that of the code and is larger than threshold $T$.

\item If no image is selected by the previous rule, select the single image with the largest activation for the matching component.
\end{enumerate}

\begin{figure}
\includegraphics[width=\linewidth]{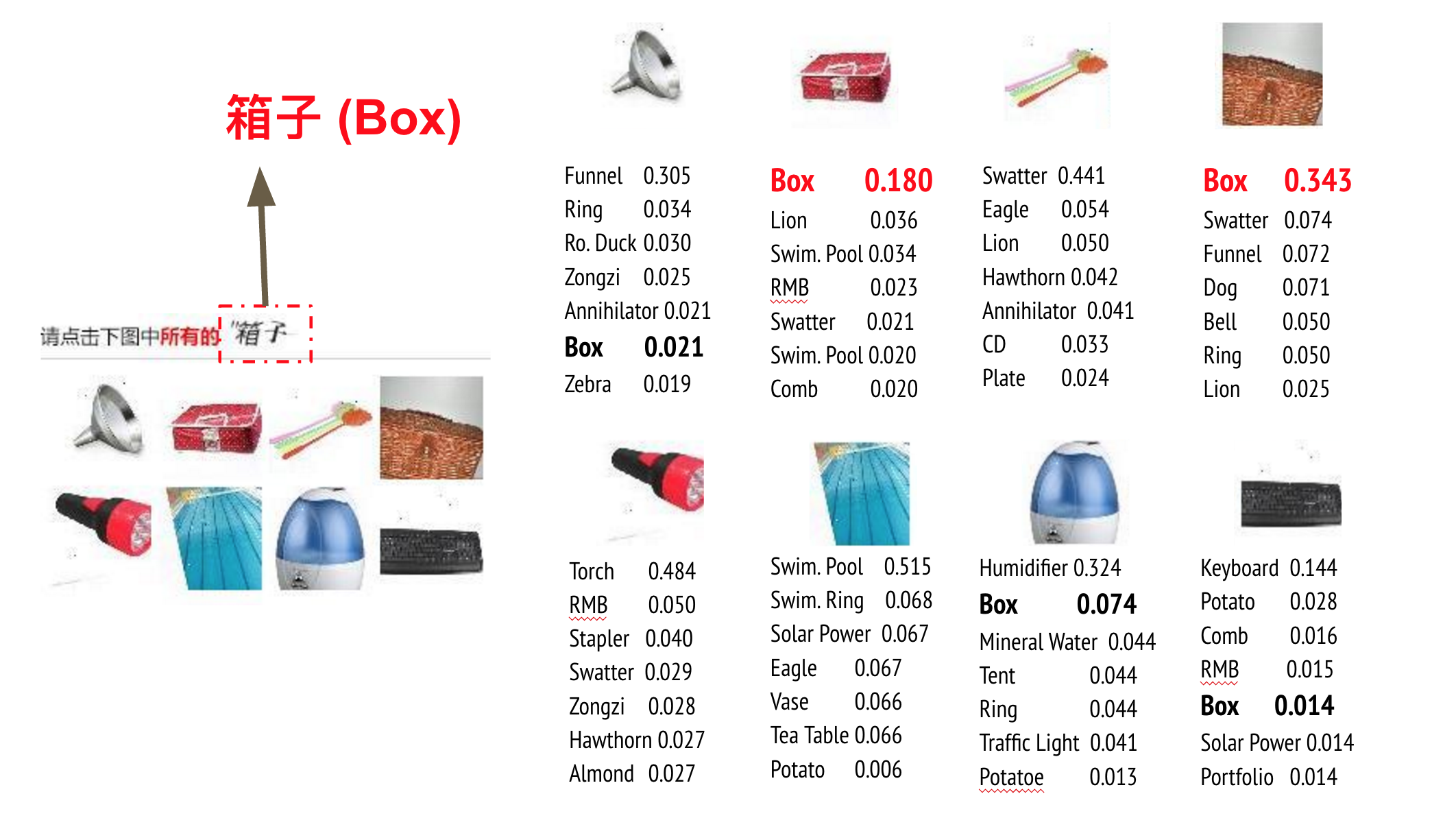}
\caption{An example of image selection. Of the 230 dimensions, the code's largest component activation was in the basis function ``box.'' The largest activations for each image are also shown. In this case, we selected the second and the fourth images. Note that ``box'' was also among the largest activations for images 7 and 8. However, they were both below the threshold, $T=0.1$.}
\label{fig:final1}
\end{figure}

For an example, see Figure \ref{fig:final1}. By applying a threshold and limiting the number of selected images, our algorithm makes good choices in the face of noisy information.

\section{Results \& Analysis}

We summarize our results below. We start by investigating the effectiveness of each of the major components of our system. We then measure the performance of the entire system in recognizing new captcha instances.

\subsection{Code Recognition Accuracy} \label{sec:codeRecognition}

We tested how well the system performs in arriving at the correct phrase of the captcha based solely on the code. By ``correct phrase'' here we mean the translation of the code presented in the captcha, from which the user has to infer the correct images. This test corresponds to testing the average accuracy of the code vertices, which will then be propagated through our association graph. We found that the accuracy of recognizing the correct Chinese phrase was 80\% on a test set of 7,000 hand-labeled codes.

\subsection{Learning from the Image Manifold}

We sought to quantify the effects of the label propagation algorithm. In other words, we wanted to understand how much information could be leveraged from the approximate image manifold. To do this, we started by testing whether the image-code co-occurrence information alone could be used to correctly identify the phrase associated with an image. We assembled a randomized set of 200 images from our database in which the correct phrase was among the Chinese phrases used to train the LeNet-5 \cite{Lecun1998Gradient} DCNN. We then tested the top $K$ image classification accuracy for those images and compared this to the top $K$ accuracy of classifying them after running label propagation. The results are shown in Table \ref{table:imageClassification}. From this table, we can see that label propagation significantly improves the prediction accuracy. The algorithm leverages the additional information provided by the graph to improve the accuracy for all images in the graph, even those with only a few occurrences, by learning from its most similar neighbors. However, you will notice that classification accuracy is low compared to other methods. This is unsurprising because the goal of our approach was not to classify the images but to learn the correct association between the codes and the images. As we will see, this is still sufficient to achieve excellent performance in defeating the captcha.

\begin{table}
\centering
\caption{Effect of Label Propagation}
\label{table:imageClassification}
\begin{tabular}{|c|c|c|}
\hline
Metric & Image-code Co-occurrence  & After Label Propagation  \\ \hline
Top 1  & 33.5\%                    & 36.5\%                   \\ \hline
Top 3  & 44.5\%                    & 49.5\%                   \\ \hline
Top 5  & 48.0\%                    & 54.5\%                   \\ \hline
\end{tabular}
\end{table}

\subsection{Defeating the Captcha}

\begin{figure}
\centering
\includegraphics[width=0.9\linewidth]{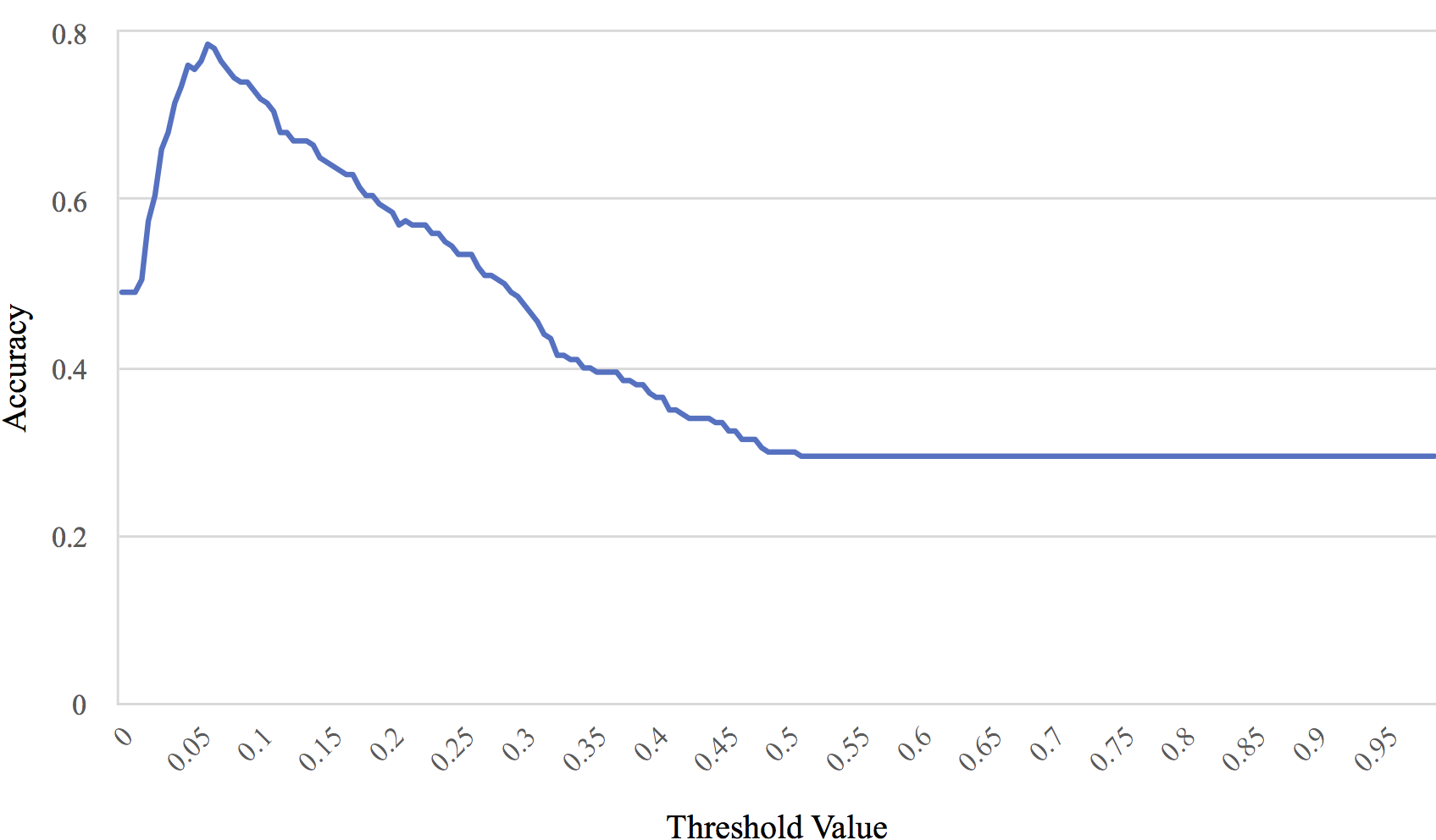}
\caption{Captcha Accuracy vs. Threshold}
\label{fig:threshold}
\end{figure}

Recall that when a captcha is presented, our system selects images based on a set of rules that rely on a threshold parameter $T$. We suspected that our system performance was dependent on a good choice for this threshold. Thus, we assembled a set of 200 new captchas as our test set and measured the accuracy of our system given various values of $T$. The results are shown in Figure \ref{fig:threshold}. With a threshold value of $T=0.65$, we found that the accuracy of our system was $154/200 = 77\%$. In recognizing a captcha, our system's response time was approximately 2 seconds, on par with human performance.

Considering average human performance is between 61\% and 98\% on image-based captchas \cite{Bursztein2010How}, these results are significant. As further evidence of this fact, our system shows promise of improvement over an approach for defeating Google's reCaptcha system published by Shet et al.\ \cite{Shet2014Are}. Their system achieves 70.78\% accuracy \cite{Sivakorn2016I}. Although direct comparisons are difficult due to differences in the underlying captcha implementations, the two captchas both require a user to select relevant images based on displayed text.

However, our approach is limited for previously unseen images. In order to get a handle on the size of this problem, we estimated the number of images that are not in our graph. We started by sampling 1,165 images from presented captchas and found that 1,123 of these were present in our graph. This means our coverage is about 96.3\%. Assuming images are shown at random, we then estimated that the true size of the image database is $3,513,055 / 0.963 \approx 3,648,032$, so we are missing approximately 134,977 images. Although the randomness assumption is likely incorrect, without it this estimation becomes much more complicated. In fact, this is an instance of the missing species problem, as described in \cite{Efron2012Large}. Since the fraction of unrepresented images was so small, we felt that simply ignoring them was an effective solution to this limitation. However, although we estimated that this effect was small, we also thought about more elegant solutions. Rather than simply ignoring these images, a better approach may be to use a $k$-nearest neighbors algorithm to find the most similar images among those in our graph. However, we would need to carefully consider the implementation of this approach so as to not significantly degrade our system's responsiveness.

\subsection{Failure Cases}

We were also interested in the cases where our approach failed. Among these cases, we found that there were two distinct types of mistakes made by our system:
\begin{enumerate}
\item Picking unrelated images when target component activations were slightly larger than the threshold.
\item Missing related images when target component activations were smaller than the threshold.
\end{enumerate}

We concluded that these two types of failures are both caused by the lack of information and significant noise. Though label propagation helps pass information through high-density areas of the manifold, significant parts of the graph are still largely uncertain. For example, in a sub-graph where many images have only been seen a few times, the noise dominates the information gained from co-occurrence statistics. One solution to this problem would be to collect more captchas in order to eliminate the uncertainty in the graph. Another solution might be to adjust the image selection rules with a dynamic threshold.

\section{Conclusion}

In this work, we develop an approach for unsupervised learning of associations between co-occurring perceptual events using a large graph. We applied this approach to successfully solve the image captcha of China's railroad system. We leveraged the structure of presented data of a particular captcha system to construct an association graph of 6 million vertices, and linked these vertices based on co-occurrence information and feature similarity between pairs of images. The resulting system, using label propagation, solved captchas with 77\% accuracy and an average response time of 2 seconds, a performance that is comparable to that of humans. The primary contributions of our work lie in system integration---exploiting and integrating existing algorithms and techniques such as deep learning, label propagation, visual hashing, and graph theory in a novel manner to create a new working system for solving a challenging technical problem in real time.

In developing this approach, we were inspired by works in multi-modal integration \cite{Srivastava2012Multimodal,Bahdanau2014Neural,Bahdanau2015End,Gregor2015DRAW,Mansimov2015Generating,Karpathy2015Deep} and association memory networks \cite{Hinton2006Fast,Mansimov2015Generating}, as well as association coding principles in biology \cite{Barlow1987Cerebral}. Barlow argued that the brain builds a statistical model of the world by learning associations between events based on their suspicious coincidence. Thus, we believe our work, by exploiting this principle, might have deeper significance beyond beating a particular captcha. We feel our system codifies a general computational framework that might be relevant for understanding association learning in the brain, specifically the mechanisms involved in learning language and associating written or phonetic words with specific objects.

There are a few important aspects of our system that resemble the problem faced by children when learning their first language. First, the relevant text in the captcha is a distorted image of Chinese characters. This distortion is like the variations in medium, size, and font of written language that a child will encounter while learning their first language. Second, distortion prevents an automated system from directly storing the textual information conveyed by the code. As a result, our system is forced to treat the codes as simply another set of visual stimuli. This is similar to how an infant, before it has learned to read, does not understand that letters convey textual information. Third, only a subset of the presented images are relevant to the presented code. This is analogous to the problem posed to children by the complex environments in which relevant objects are always in the company of distractors. Therefore, a child has to discover what is relevant by observing and learning from coincident events. Finally, neither the phrases nor the presented images are equipped with categorical information. This resembles the unsupervised setting in which children gradually discover semantic meaning of words through their association with observed objects and events. Thus, this class of large-scale association graphs and related graph algorithms, such as label and belief propagation, could be useful for conceptualizing the mechanisms that enable us to discover the semantic meanings of words in an unsupervised manner.

\section*{Acknowledgments}

The authors would like to thank Professor Christos Faloustos, Professor William Cohen, Jingyuan Liu, Yuxin Chen, Yimeng Zhang, Wenhao Zhang, Qin Lyu, and Tong He for their help and support in this project.

This work was supported by NSF grant CISE 1320651 and by IARPA via DOI contract D16PC00007. The U.S. Government is authorized to reproduce and distribute reprints for governmental purposes notwithstanding any copyright annotation thereon. However, the views and conclusions contained within should not be interpreted as necessarily representing the official policies or endorsements of the funding agencies.

\bibliographystyle{IEEEtran}
\bibliography{reference.bib}

\begin{thebibliography}{10}
\providecommand{\url}[1]{#1}
\csname url@samestyle\endcsname
\providecommand{\newblock}{\relax}
\providecommand{\bibinfo}[2]{#2}
\providecommand{\BIBentrySTDinterwordspacing}{\spaceskip=0pt\relax}
\providecommand{\BIBentryALTinterwordstretchfactor}{4}
\providecommand{\BIBentryALTinterwordspacing}{\spaceskip=\fontdimen2\font plus
\BIBentryALTinterwordstretchfactor\fontdimen3\font minus
  \fontdimen4\font\relax}
\providecommand{\BIBforeignlanguage}[2]{{%
\expandafter\ifx\csname l@#1\endcsname\relax
\typeout{** WARNING: IEEEtran.bst: No hyphenation pattern has been}%
\typeout{** loaded for the language `#1'. Using the pattern for}%
\typeout{** the default language instead.}%
\else
\language=\csname l@#1\endcsname
\fi
#2}}
\providecommand{\BIBdecl}{\relax}
\BIBdecl

\bibitem{Barlow1987Cerebral}
H.~Barlow, \emph{Cerebral Cortex as Model Builder}, 1987.

\bibitem{Zhu2016China}
\BIBentryALTinterwordspacing
J.~Zhu, ``{China Railway Corporation: Scaling Online Sales for the Largest
  Railway in the World},'' Pivotal, 2016. [Online]. Available:
  \url{https://content.pivotal.io/case-studies/china-railway-corporation}
\BIBentrySTDinterwordspacing

\bibitem{Shet2014Are}
\BIBentryALTinterwordspacing
V.~Shet, ``{A}re you a robot? {I}ntroducing {``No CAPTCHA reCAPTCHA''},''
  Google, 2014. [Online]. Available:
  \url{https://security.googleblog.com/2014/12/are-you-robot-introducing-no-captcha.html}
\BIBentrySTDinterwordspacing

\bibitem{Krizhevsky2012ImageNet}
A.~Krizhevsky, I.~Sutskever, and G.~E. Hinton, ``Imagenet classification with
  deep convolutional neural networks,'' in \emph{Proceedings of Advances in
  Neural Information Processing Systems 25}, 2012.

\bibitem{Zhu2003Semi}
X.~Zhu, Z.~Ghahramani, and J.~Lafferty, ``Semi-supervised learning using
  gaussian fields and harmonic functions,'' in \emph{Proceedings of the 20th
  International Conference on Machine Learning}, 2003.

\bibitem{Bursztein2010How}
E.~Bursztein, S.~Bethard, C.~Fabry, J.~C. Mitchell, and D.~Jurafsky, ``How good
  are humans at solving {CAPTCHAs?} {A} large scale evaluation,'' in
  \emph{Proceedings of the 2010 {IEEE} Symposium on Security and Privacy},
  2010.

\bibitem{Sivakorn2016I}
S.~Sivakorn, I.~Polakis, and A.~D. Keromytis, ``I am robot: ({D}eep) learning
  to break semantic image {CAPTCHAs},'' in \emph{Proceedings of the 2016 {IEEE}
  European Symposium on Security and Privacy}, 2016.

\bibitem{Ahn2003CAPTCHA}
L.~V. Ahn, M.~Blum, N.~J. Hopper, and J.~Langford, ``{CAPTCHA:} {U}sing hard
  {AI} problems for security,'' in \emph{Proceedings of the 22nd International
  Conference on Theory and Applications of Cryptographic Techniques}, 2003.

\bibitem{Goodfellow2013Multi}
I.~J. Goodfellow, Y.~Bulatov, J.~Ibarz, S.~Arnoud, and V.~D. Shet,
  ``Multi-digit number recognition from street view imagery using deep
  convolutional neural networks,'' \emph{CoRR}, 2013.

\bibitem{2013Vicarious}
\BIBentryALTinterwordspacing
``Vicarious {AI} passes first {T}uring test: {CAPTCHA},'' Vicarious, 2013.
  [Online]. Available:
  \url{http://news.vicarious.com/post/65316134613/vicarious-ai-passes-first-turing-test-captcha}
\BIBentrySTDinterwordspacing

\bibitem{Bursztein2014End}
E.~Bursztein, J.~Aigrain, A.~Moscicki, and J.~C. Mitchell, ``The end is nigh:
  {G}eneric solving of text-based {CAPTCHAs},'' in \emph{Proceedings of the 8th
  USENIX Workshop on Offensive Technologies}, San Diego, CA, 2014.

\bibitem{Karpathy2014Large}
A.~Karpathy, G.~Toderici, S.~Shetty, T.~Leung, R.~Sukthankar, and L.~Fei-Fei,
  ``Large-scale video classification with convolutional neural networks,'' in
  \emph{Proceedings of the 2014 {IEEE} Conference on Computer Vision and
  Pattern Recognition}, 2014.

\bibitem{Simonyan2014Two}
K.~Simonyan and A.~Zisserman, ``Two-stream convolutional networks for action
  recognition in videos,'' in \emph{Proceedings of Advances in Neural
  Information Processing Systems 27}, 2014.

\bibitem{Razavian2014CNN}
A.~S. Razavian, H.~Azizpour, J.~Sullivan, and S.~Carlsson, ``{CNN} features
  off-the-shelf: {A}n astounding baseline for recognition,'' in
  \emph{Proceedings of the 2014 {IEEE} Conference on Computer Vision and
  Pattern Recognition Workshops}, 2014.

\bibitem{Lin2016Transfer}
X.~Lin, H.~Wang, Z.~Li, Y.~Zhang, A.~Yuille, and T.~S. Lee, ``Transfer of
  view-manifold learning to similarity perception of novel objects,'' 2016.

\bibitem{Bengio2013Using}
S.~Bengio, J.~Dean, D.~Erhan, E.~Ie, Q.~V. Le, A.~Rabinovich, J.~Shlens, and
  Y.~Singer, ``Using web co-occurrence statistics for improving image
  categorization,'' \emph{CoRR}, 2013.

\bibitem{Ising1925Beitrag}
E.~Ising, ``Beitrag zur theorie des ferromagnetismus,'' \emph{Zeitschrift
  f{\"u}r Physik}, 1925.

\bibitem{Zhu2005Semi}
X.~Zhu, ``Semi-supervised learning with graphs,'' 2005.

\bibitem{Hinton2006Fast}
G.~E. Hinton, S.~Osindero, and Y.-W. Teh, ``A fast learning algorithm for deep
  belief nets,'' \emph{Neural Computation}, 2006.

\bibitem{Mansimov2015Generating}
E.~Mansimov, E.~Parisotto, L.~J. Ba, and R.~Salakhutdinov, ``Generating images
  from captions with attention,'' \emph{CoRR}, 2015.

\bibitem{Lecun1998Gradient}
Y.~LeCun, L.~Bottou, Y.~Bengio, and P.~Haffner, ``Gradient-based learning
  applied to document recognition,'' \emph{Proceedings of the {IEEE}}, 1998.

\bibitem{Klinger2010pHash}
\BIBentryALTinterwordspacing
E.~Klinger and D.~Starkweather, ``{pHash:} {T}he open source perceptual hash
  library,'' 2010. [Online]. Available: \url{http://www.phash.org}
\BIBentrySTDinterwordspacing

\bibitem{Hu2015Transferring}
F.~Hu, G.-S. Xia, J.~Hu, and L.~Zhang, ``Transferring deep convolutional neural
  networks for the scene classification of high-resolution remote sensing
  imagery,'' \emph{Remote Sensing}, 2015.

\bibitem{Bahdanau2014Neural}
D.~Bahdanau, K.~Cho, and Y.~Bengio, ``Neural machine translation by jointly
  learning to align and translate,'' \emph{CoRR}, 2014.

\bibitem{Bahdanau2015End}
D.~Bahdanau, J.~Chorowski, D.~Serdyuk, P.~Brakel, and Y.~Bengio, ``End-to-end
  attention-based large vocabulary speech recognition,'' \emph{CoRR}, 2015.

\bibitem{Gregor2015DRAW}
K.~Gregor, I.~Danihelka, A.~Graves, and D.~Wierstra, ``{DRAW:} {A} recurrent
  neural network for image generation,'' \emph{CoRR}, 2015.

\bibitem{Karpathy2015Deep}
A.~Karpathy and L.~Fei-Fei, ``Deep visual-semantic alignments for generating
  image descriptions,'' in \emph{Proceedings of he 2015 {IEEE} Conference on
  Computer Vision and Pattern Recognition}, 2015.

\bibitem{Efron2012Large}
B.~Efron, \emph{Large-Scale Inference}, 2012.

\bibitem{Srivastava2012Multimodal}
N.~Srivastava and R.~Salakhutdinov, ``Multimodal learning with deep boltzmann
  machines,'' in \emph{Proceedings of Advances in Neural Information Processing
  Systems 25}, 2012.

\end{thebibliography}

\end{document}